\documentclass[journal]{IEEEtran}

\usepackage{xcolor,soul,framed} 

\colorlet{shadecolor}{yellow}
\usepackage[pdftex]{graphicx}
\graphicspath{{../pdf/}{../jpeg/}}
\DeclareGraphicsExtensions{.pdf,.jpeg,.png}

\usepackage[cmex10]{amsmath}
\usepackage{array}
\usepackage{mdwmath}
\usepackage{mdwtab}
\usepackage{eqparbox}
\usepackage{url}

\usepackage{booktabs}
\usepackage{subfigure}
\usepackage{bbding}
\usepackage{multirow}
\usepackage{tabularx}
\usepackage{amssymb}

\usepackage{hyperref}
\usepackage[capitalize]{cleveref}
\crefname{section}{Sec.}{Secs.}
\Crefname{section}{Section}{Sections}
\Crefname{table}{Table}{Tables}
\crefname{table}{Table}{Tables}

\begin{document}
\bstctlcite{IEEEexample:BSTcontrol}
    \title{Disentangled Contrastive Image Translation for Nighttime Surveillance}
  \author{
        Guanzhou Lan,
        Bin Zhao,
        Xuelong Li,~\IEEEmembership{Fellow,~IEEE}

  }

\markboth{IEEE TRANSACTIONS ON Image Processing
}{Roberg \MakeLowercase{\textit{et al.}}: Disentangled Contrastive Image Translation for Nighttime Surveillance}

\maketitle

\begin{abstract}
Nighttime surveillance suffers from degradation due to poor illumination and arduous human annotations. It is challengable and remains a security risk at night. Existing methods rely on multi-spectral images to perceive objects in the dark, which are troubled by low resolution and color absence. 
We argue that the ultimate solution for nighttime surveillance is night-to-day translation, or \emph{Night2Day}, which aims to translate a surveillance scene from nighttime to the daytime while maintaining semantic consistency. To achieve this, this paper presents a Disentangled Contrastive ($\emph{DiCo}$) learning method. Specifically, to address the poor and complex illumination in the nighttime scenes, we propose a learnable physical prior, \emph{i.e.}, the color invariant, which provides a stable perception of a highly dynamic night environment and can be incorporated into the learning pipeline of neural networks. Targeting the surveillance scenes, we develop a disentangled representation, which is an auxiliary pretext task that separates surveillance scenes into the foreground and background with contrastive learning. Such a strategy can extract the semantics without supervision and boost our model to achieve instance-aware translation.
Finally, we incorporate all the modules above into generative adversarial networks and achieve high-fidelity translation. 
This paper also contributes a new surveillance dataset called NightSuR. It includes six scenes to support the study on nighttime surveillance. This dataset collects nighttime images with different properties of nighttime environments, such as flare and extreme darkness.
 Extensive experiments demonstrate that our method outperforms existing works significantly. The dataset and source code will be released on GitHub soon.
\end{abstract}


\begin{IEEEkeywords}
nighttime vision, image translation, disentanged representation, contrastive learning
\end{IEEEkeywords}

%
\IEEEpeerreviewmaketitle


\section{Introduction}\label{Sec: Intro}

\IEEEPARstart{L}{ow}-light environment prompts human to use rod photoreceptors rather than cone photoreceptors operating in daytime \cite{gegenfurtner1999seeing}. It results in poor spatial resolution, diminished contrast sensitivity, and absent color vision at night. Similar problems appear in imaging systems, shown as low lightness, contrast, and resolution with high ISO noise \cite{Zheng_2021_ICCV}. Those issues challenge the visual perception and cognition of surveillance in the dark, which results in security threats in the night environment.

To overcome such difficulty, plenty of efforts have been made. One option is multi-spectral image analysis, such as the fusion of visible and infrared images\cite{tang2023divfusion, li2020mdlatlrr}. Nevertheless, those techniques depend on adequate multisource information retrieval and result in resolution and color destruction frequently. Another option is to improve the clarity and details of images using low-light image enhancement \cite{ren2019low, jiang2021enlightengan, Lamba_2023_WACV}. However, it cannot deal with pixel-wise information loss and domain invariance from nighttime to daytime images.

\begin{figure}[t]
\centering
\includegraphics[width=1.0\linewidth]{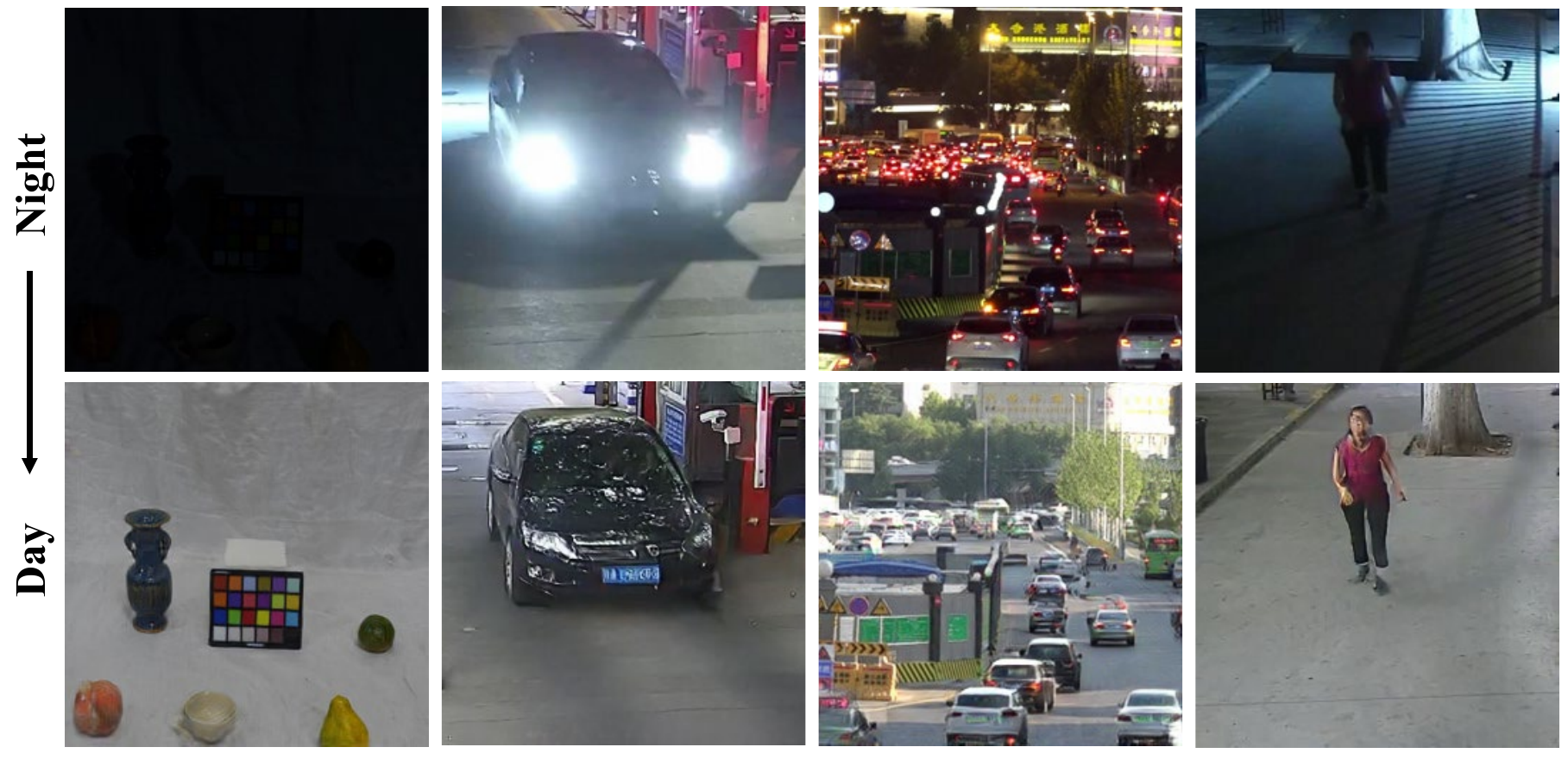} \
\vspace{-0.5cm}
\caption{Some samples of nighttime surveillance. The first row are the night images, and the second are the generated images by the proposed method.}
\label{fig:top}
\vspace{-0.4cm}
\end{figure}

\begin{figure*}[t]
\centering
\includegraphics[width=1.0\linewidth]{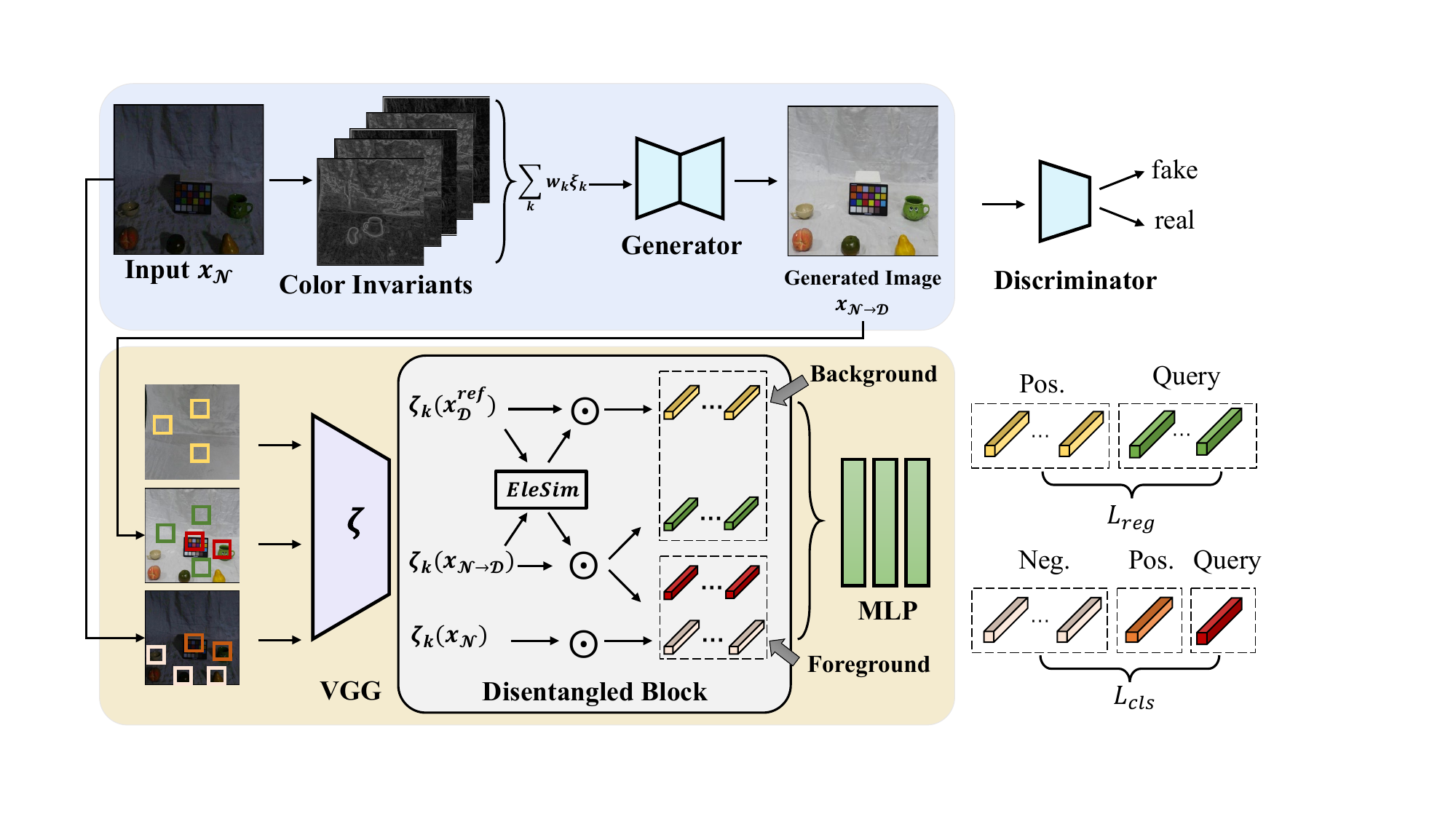} \
\vspace{-0.2cm}
\caption{The overall architecture of DiCo. The test stage is the first row in the blue block. The disentangled representation is the second row in the yellow block. The contrastive learning strategy is presented at the end of the second row.}
\vspace{-0.4cm}
\label{fig:one}
\end{figure*}

We argue that the ultimate solution for nighttime surveillance is night-to-day translation, or \emph{Night2Day}, which aims to improve night scene perceptual quality by transforming an image from night to day while maintaining semantic consistency of image content in a generative way. It consequently restores the pixel-wise information loss and eliminates the domain gap between nighttime and daytime images, which not only comforts human eyes but also benefits the subsequent tasks, such as object detection and instance segmentation. It is challenging for the following significant properties: (1) Nighttime images are more difficult to represent for the widespread distribution shift in nighttime images. \emph{e.g.}, the extreme dimming light and flare both exist in the night domain and even in a single image. (2) The supervision is not accessible because it is impossible to obtain the paired nighttime and daytime images in the real world. (3) Weak supervision from human annotations is also not accessible. It is usually out of human cognition at night, especially in surveillance cases. These issues influence learning target distribution or keep the original semantics less or more.



In this paper, we present an unsupervised method for real-world nighttime surveillance, dubbed as $\bf{Di}$sentangled $\bf{Co}$ntrastive learning ($\bf{DiCo}$).\cref{fig:one} shows the overview of $\bf{DiCo}$. $\bf{DiCo}$ tries to deal with the issues discussed above with two intuitions: (1) Employing the physical prior to reduce the distribution shift in the night; (2) Fully utilizing the fixed scene prior in surveillance scenes as the supervision. Firstly, derived from the Kubelka-Munk theory, a learnable color invariant is designed to access stable perception in the highly dynamic nighttime environment. Secondly, a disentanglement module is developed to separate images into foreground and background. This module offers a disentangled representation of nighttime and daytime images with only the reference to the images in the daytime. Thirdly, we design a disentangled contrastive learning strategy for the foreground and background to keep the semantic consistency. 
Finally, all the modules above are incorporated into the generative adversarial network to learn the target distribution.
Moreover, we also contribute a dataset with various surveillance scenes for Night2Day.
Our main contributions lie in the following three folds:

1. To present the highly dynamic nighttime environment, we propose a stable perception strategy with the proposed learnable color invariant. It reduces the distribution shift of the model caused by the extreme light conditions in nighttime surveillance.

2.	Targeting the surveillance scenes, we propose a novel disentangled representation for foreground and background features referring to the daytime domain. In this case, better visual effects and instance-aware image translation are achieved in unsupervised conditions.

3. We also publish a new dataset for nighttime surveillance scenes called \emph{NightSuR}. This dataset includes six scenes and 6574 images with extreme dimming and flare cases.

\section{Related Work}

As aforementioned, nighttime surveillance can be achieved by low-light image enhancement, image-to-image translation, and Night2Day. In the following, the three tasks are reviewed and discussed in detail. 

Low-light image enhancement is a traditional low-level task in the night scene. Guo et al. propose a traditional method to estimate the illumination map by exposing the structure prior \cite{guo2016lime}. Li et al. propose a robust Retinex model and develop an optimization approach \cite{li2018structure}.
Recently, as deep learning develops, more low-light image enhancement solutions are designed based on deep neural networks. Ren et al. propose a novel spatially variant recurrent neural network to compose an encoder-decoder model \cite{ren2019low}. Furthermore, Xu et al. apply the attention module to enhance the low-light image in decomposition \cite{xu2020learning}. Guo et al. expand the generalization of the deep learning model on low light image enhancement \cite{guo2020zero}. Besides, EnlightenGAN employs a pretrained VGG and adversarial training which achieves excellent results \cite{jiang2021enlightengan}. By employing the Retinex model, URetinex-Net achieves new state-of-the-art enhancement\cite{Wu_2022_CVPR}. Another trend is: researchers prefer to distinguish between nighttime enhancement and low-light enhancement due to the complex light condition of nighttime environments \cite{sharma2021nighttime, jin2022unsupervised}.

The image-to-image translation is proposed to synthesize new filtered target images \cite{hertzmann2001image}.  
Since generative adversarial networks show competitive performance in image generation \cite{goodfellow2014generative}, lots of image-to-image translation models have been proposed and show impressive results \cite{lee2018diverse, 9694500, Ntavelis_2022_CVPR}. As the diffusion model developed, a new research paradigm for image-to-image translation is also proposed. One of the representative works is Palette \cite{saharia2022palette}.

Particularly, for unpaired image translation, cycle-consistency training \cite{zhu2017unpaired} has become a strong baseline and has been applied in extensive later works \cite{anokhin2020high, Han_2021_CVPR, xu2022maximum}.
Recently, one-side translation without cycle-consistency training raises the attention of researchers. Park et al. \cite{park2020contrastive} first introduce contrastive learning to image-to-image translation and get excellent performance. \cite{liu2021divco} apply contrastive learning into a conditional GAN. \cite{zheng2021spatially} perform patch-wise contrastive learning on self-similarity maps and present a strong ability on controlling the semantic consistency of generated images. \cite{wang2021instance} employ a generative strategy to mine hard negative examples during contrastive learning. Improving the contrastive learning strategy, a batch of methods reach new state-of-the-art in unpaired image translation\cite{Jung_2022_CVPR, hu2022qs, Zhan_2022_CVPR}.

Night2Day towards driving scenes has received much attention. \cite{anoosheh2019night} first propose ToDayGAN for the Night2Day task. \cite{zheng2020forkgan} propose ForkGAN to improve visual perceptual quality on a rainy night. Some works also discuss Night2Day briefly. Inspired by InstaGAN \cite{mo2018instagan}, some following works change Night2Day into supervised settings to improve the results in driving scenes. \cite{shen2019towards} contribute a dataset in 4 domains with sufficient annotations, including daytime and nighttime images, and develop a supervised method to translate images across domains. Following INIT \cite{shen2019towards}, many instance-aware translation works have been proposed for driving scenes in supervised settings \cite{bhattacharjee2020dunit, jeong2021memory, kim2022instaformer}.




\begin{figure*}[t]
\centering
\includegraphics[width=0.9\linewidth]{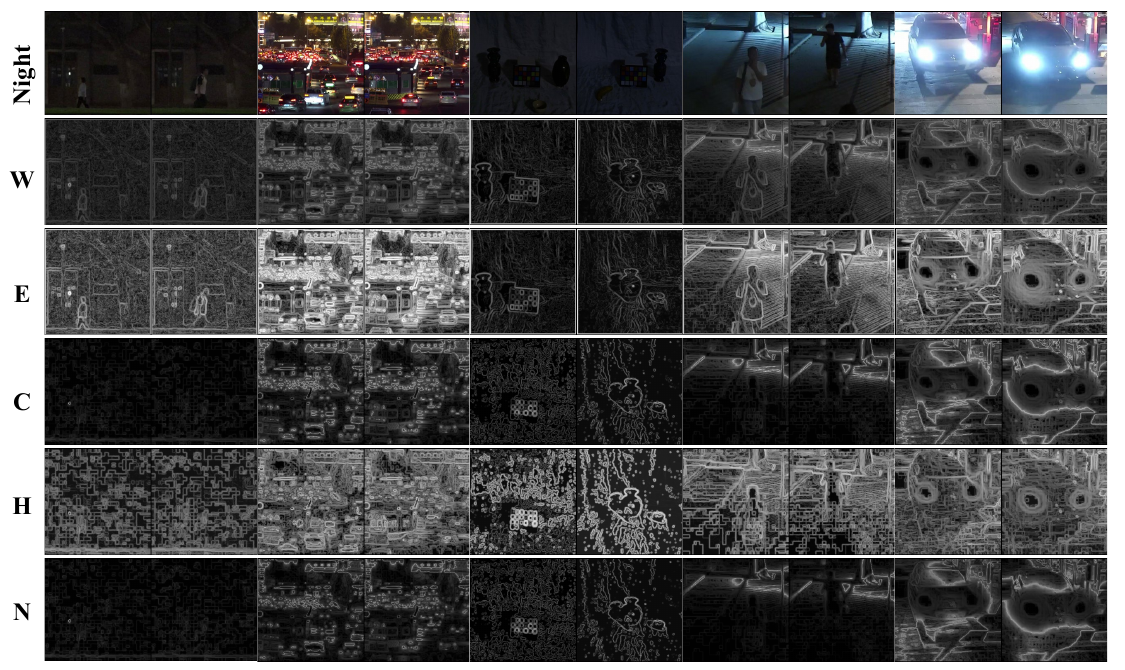}
\vspace{-0.4cm}
\caption{The visualization of different color invariants in different scenes.}
\label{fig:IC}
\end{figure*} 

\begin{table*}[htbp]

\centering
\renewcommand\arraystretch{1.5}
\caption{Different color invariants explored by \cite{geusebroek2001color}. The SG, FR, II, and IC denote the invariance properties to $\bf{S}$cene $\bf{G}$eometry, $\bf{F}$resnel $\bf{R}$eflections, $\bf{I}$llumination $\bf{I}$ntensity, and $\bf{I}$llumination $\bf{C}$olor. }
\begin{tabular}{@{}lccccc@{}}
    \toprule
    Invariant	 & Difinition  & SG   & FR   & II & IC 	\\
    \hline
    
    $E$ \ & $E = \sqrt{E_i^2 + E_{\lambda i}^2 + E_{\lambda \lambda i}^2 + E_{j}^2 + E_{\lambda j}^2 + E_{\lambda \lambda j}^2} $ & \XSolidBrush &\XSolidBrush & \XSolidBrush&\XSolidBrush \\
    
    \multirow{2}{*}{$W$} & $W = \sqrt{W_i^2 + W_{\lambda i}^2 + W_{\lambda \lambda i}^2 + W_{j}^2 + W_{\lambda j}^2 + W_{\lambda \lambda j}^2}$&\multirow{2}{*}{ \XSolidBrush}&\multirow{2}{*}{ \XSolidBrush}&\multirow{2}{*}{ \Checkmark}&\multirow{2}{*}{ \XSolidBrush} \\ &$W_i = \frac{E_i}{E}$, $W_{\lambda i} = \frac{E_{\lambda i}}{E}$, $W_{\lambda \lambda i} = \frac{E_{\lambda \lambda i}}{E}$ 
    \\
    
    \multirow{2}{*}{$C$} & $C= \sqrt{C_{\lambda i}^2 + C_{\lambda \lambda i}^2 + C_{\lambda j}^2 + C_{\lambda \lambda j}^2}$  &\multirow{2}{*}{ \Checkmark}&\multirow{2}{*}{ \XSolidBrush}&\multirow{2}{*}{ \Checkmark}&\multirow{2}{*}{ \XSolidBrush}\\ &$C_{\lambda i} = \frac{E_ {\lambda i}E-E_\lambda E_i}{E^2}$, $C_{\lambda \lambda i} = \frac{E_ {\lambda \lambda i}E-E_{\lambda \lambda} E_i}{E^2}$
    \\
    
    $H$ & $H= \sqrt{H_{i}^2 + H_{j}^2}$ , $H_{ i} = \frac{E_ {\lambda i}E{\lambda \lambda}-E_\lambda E_{\lambda \lambda i}}{E_\lambda^2+E_{\lambda \lambda}^2}$
    \
    & \Checkmark& \Checkmark& \Checkmark& \XSolidBrush \\
    
        \multirow{2}{*}{$N$} & $N= \sqrt{N_{\lambda i}^2 + N_{\lambda \lambda i}^2 + N_{\lambda j}^2 + N_{\lambda \lambda j}^2}$  &\multirow{2}{*}{ \Checkmark}&\multirow{2}{*}{ \XSolidBrush}&\multirow{2}{*}{ \Checkmark}&\multirow{2}{*}{ \Checkmark}\\ 
    
    &$N_{\lambda i} = \frac{E_ {\lambda i}E-E_\lambda E_i}{E^2}$, $N_{\lambda \lambda i} = \frac{E_ {\lambda \lambda i}E^2-E_{\lambda \lambda} E_iE - 2E_{\lambda i}E_\lambda E + 2 E_\lambda^2E_i}{E^3}$    \\
    
    \bottomrule

\end{tabular}
\label{tab:CI}
\end{table*}

\section{Disentangled Contrastive Networks} \label{Sec: DCN}
Given nighttime image $x_\mathcal{N} \in \mathcal{N}$ and daytime image $x_\mathcal{D} \in \mathcal{D}$, Night2Day aims to translate images from nighttime to daytime while maintaining content semantic consistency. It needs to construct a map $\mathcal{F}$ with parameters $\mathcal{W}$, which can be modeled as:
\begin{equation}
    \begin{aligned}
        \ & \mathcal{F}_\mathcal{W}: \mathcal{N} \rightarrow \mathcal{D}. \\
    \end{aligned}	
    \label{eq: task formulation}
\end{equation}

As the issues discussed in the introduction, it is difficult to set the optimization target to obtain the parameters $\mathcal{W}$. However, in surveillance scenes, the fixed scenes provide supervision in the background regions of the image. If we can obtain the disentanglement of the background and foreground, the optimization target can be modeled as:
\begin{equation}
    \begin{aligned}
        \ & \min d_{back} \,(\zeta^b(x^{ref}_\mathcal{D}), \zeta^b(x_\mathcal{N\rightarrow D})) \\
        \ & + d_{fore} \, (\zeta^f(x_\mathcal{N}), \zeta^f(x_\mathcal{N\rightarrow D})),
    \end{aligned}	
    \label{eq: optimization target}
\end{equation}
where $d_{back}$ and $d_{fore}$ are distance measurements. $\zeta^f$ and $\zeta^b$ denote the features of foreground and background.
$d_{back}$ can employ the L1/L2 norm to get direct supervision from $x^{ref}_\mathcal{D}$. Despite $d_{fore}$ still needs to carefully design to prevent the domination of $x_\mathcal{N}$, this modeling has greatly reduce the supervision dilemma discussed before.

In this section, We will introduce our Disentangled Contrastive Networks in detail. First, we introduce the proposed learnable color invariant, which provides robust perception in flare and darkness. Then we present our disentangled representation of the background and foreground based on the element-wise Pearson correlation coefficient. Finally, we incorporate all the modules into the disentangled contrastive learning framework to achieve semantic preserving \emph{Night2Day} in surveillance scenes.

\subsection{Learnable Color Invariant}

\begin{figure*}[t]
\centering
\includegraphics[width=1.0\linewidth]{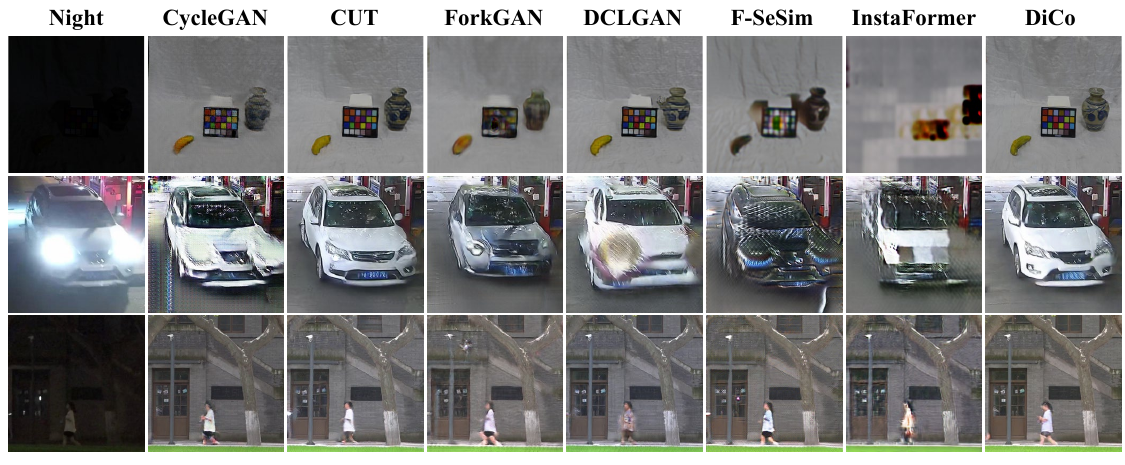} \
\vspace{-0.5cm}
\caption{The qualitative evaluation on \emph{NightSuR}. From top to down are: \textit{Darkness}, \textit{Vehicle}, and \textit{Pedestrian-b}.}
\vspace{-0.2cm}
\label{fig:results1}
\end{figure*}

The flare and darkness outside of human recognition will also result in an unstable perception of the Convolutional Neural Network (CNN), which manifests as a 
 distribution shift of feature map activations of CNN's layers and ultimately leads to the collapse of the flare and darkness regions of synthetic images, as depicted in \cref{fig:results1}. 
Fortunately, color invariants can represent object properties regardless of the recording conditions, especially the low illumination. It motivates us to search for a color invariant that is resistant to flare and darkness and has stable perception \cite{geusebroek2001color}.

Searching for color invariants typically relies on a photometric model. Our learnable color invariant employs Geusebroek's invariant edge detectors \cite{geusebroek2001color}, which is derived from Kubelka-Munk theory \cite{kubelka1931beitrag}. It describes the spectrum of light $E$ reflected from an object in the view direction as:
\begin{equation}
    \begin{aligned}
        \ & E(\lambda, \Vec{\bf{x}}) = e(\lambda, \Vec{\bf{x}})((1-\rho_f(\Vec{\bf{x}}))^2R_{\infty}(\lambda, \Vec{\bf{x}}) + \rho_f(\Vec{\bf{x}})),
    \end{aligned}	
    \label{eq: K-M}
\end{equation}
where $\Vec{\bf{x}}$ is a vector that denotes the spatial location on the image plane, and $\lambda$ is the wavelength of light. $e(\lambda, \Vec{\bf{x}})$ is the spectrum. $R_{\infty}$ represents the material reflectivity and $\rho_f$ is the Fresnel reflectance coefficient. Simplifying assumptions in \cref{eq: K-M}, the derived invariants $E$, $W$, $C$, $N$, and $H$ present edge detectors invariant to various combinations of illumination changes, including scene geometry, Fresnel reflections, and the intensity and color of the illumination. These invariant properties are the base of our learnable color invariant due to their significance for nighttime surveillance.

We introduce the invariant $E$ in the following while leaving the other invariants to the \cref{tab:CI}. The visualizations of each invariant in different illumination conditions are shown in \cref{fig:IC}.
Firstly, the Gaussian color model is employed to estimate $E$ and the partial derivatives $E_\lambda$, $E_{\lambda \lambda}$.

\begin{equation}
    \begin{aligned}
        \begin{bmatrix}
            E(i,j) \\
            E_\lambda(i,j)  \\
            E_{\lambda \lambda}(i,j)
        \end{bmatrix} = 
        \begin{bmatrix}
            &0.06,&0.63, &0.27 \\
            &0.3, &0.04, &-0.35 \\
            &0.34, &-0.6, &0.17
        \end{bmatrix}
        \begin{bmatrix}
            R(i,j) \\
            G(i,j) \\
            B(i,j) \\
        \end{bmatrix},
    \end{aligned}	
    \label{eq: Guassian}
\end{equation}
where $i,j$ are pixel locations of the image. The spatial derivatives $E_i$ and $E_j$ are calculated by convolving $E$ with Gaussian derivative kernel $g$ and standard deviation $\sigma$:
\begin{equation}
    \begin{aligned}
        \ & E_i(i, j, \sigma) = \sum_{t \in \mathbf{Z} }{E(t,j)\frac{\partial g(i-t, \sigma)}{\partial i}}.
    \end{aligned}	
    \label{eq: E_x}
\end{equation}

And the spatial derivatives for $E_{\lambda i}$ and $E_{\lambda \lambda i}$ are the same to operate \cref{eq: E_x} on the estimated $E_\lambda(i,j)$ and$ E_{\lambda \lambda}(i,j)$. Following \cite{geusebroek2001color}, the preliminary color invariant $E$ is computed as:
\begin{equation}
    \begin{aligned}
    & E = \sqrt{E_i^2 + E_{\lambda i}^2 + E_{\lambda \lambda i}^2 + E_{j}^2 + E_{\lambda j}^2 + E_{\lambda \lambda j}^2}.  \\
    \end{aligned}	
    \label{eq: E}
\end{equation}

The invariants are derived under different assumptions, which means each invariant can only provide robustness toward some properties. It is difficult to represent such a complex scene using just one invariant. Such observation is also confirmed by the results in \cref{fig:IC}. This inspires us to design the invariant adaptively.
We set a learnable parameters $\Lambda$ under ensemble learning strategy:
\begin{equation}
    \begin{aligned}
        \ & \xi = \Lambda\Phi,
    \end{aligned}	
    \label{eq: ensemble}
\end{equation}
where $\Phi$ denotes the concatenation of color invariants $E$, $W$, $C$, $N$, and $H$. $\xi$ is the final learnable color invariant.
Such design brings two benefits for night scene perception:
(1) It can adaptively adjust the weight towards the different illumination conditions, which provides more robust and flexible color invariant representations.
(2) With more observation from different invariants, it can provide more information in the input and avoid solving the ill-pose problem with one channel observation of the input and three channel variables of the generated image in Night2Day.
Later, the learnable color invariant (LCI) $\xi$ will be put into the ResNet-based generator to estimate the daytime images with the implicit model and described in the following sections.

\begin{figure*}[t]
\centering
\includegraphics[width=1.0\linewidth]{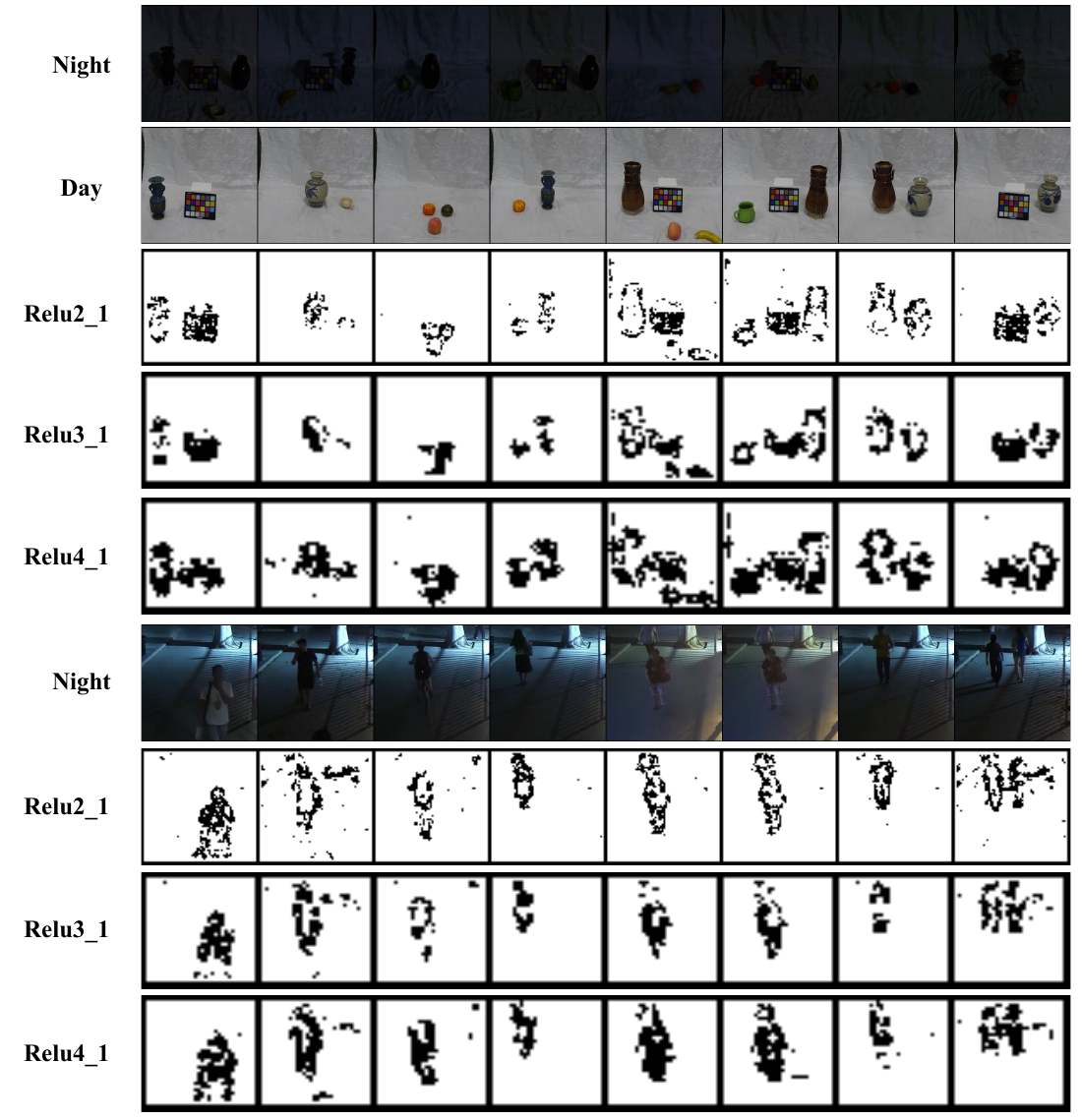}
\caption{The visualization of the disentangled representations in different layers of VGG-16.}
\vspace{-0.4cm}
\label{fig: dis}
\end{figure*}

\subsection{Element-wise Similarity} \label{Sec:Element-wise Similarity}

The disentangled representation for surveillance scenes aims to estimate a mask $M$, so as to obtain the foreground features $\zeta^f(x)$ and background features $\zeta^b(x)$ of any image $x$ as follows:
\begin{equation}
    \begin{aligned}
        \ & \zeta^b(x) = M \odot \zeta(x), \\
        & \zeta^f(x) = (1-M) \odot \zeta(x),
    \end{aligned}	
    \label{eq: disentanglement}
\end{equation}
Referring to the background features $\zeta(x^{ref})$, computing the similarity between $\zeta(x^{ref})$ and daytime image $\zeta(x_\mathcal{D})$ can provide natural clusters into background and foreground. Based on this intuition, we introduce our Element-wise Similarity ($EleSim$).

As a common practice, we employ a pretrained VGG-16 Network as the feature extractor $\zeta$. Employing the Pearson correlation coefficient as the similarity measurement, we compute the feature-level element-wise similarity between $\zeta(x_\mathcal{D})$ and the referring background $\zeta(x^{ref})$. 

Denoting features extracted from the $k-th$ layer of VGG-16 as $\zeta_k$, the size of $\zeta_k$ is $(Bs, C_k, H_k, W_k)$. 
The elements are selected from the $\zeta_k$ with the size of $(Bs, C_k, 1, 1)$.
Computing the similarity between elements from $\zeta(x_\mathcal{D})$ and $\zeta_k(x^{ref})$ in the corresponding location, we can obtain similarity scores for each element with the size of $(Bs, 1, H_k, W_k)$. The formulation of $EleSim$ is derived from the Pearson correlation coefficient:
\begin{equation} \footnotesize
    \begin{aligned}
        & EleSim(\zeta_k(i,j), \zeta_k^{ref}(i,j)) = \\
        & \dfrac{\sum (\zeta_k(i, j)-\mu_{\zeta_k}(i,j))(\zeta_k^{ref}(i, j)-\mu_{\zeta_k}(i,j))}{\sqrt{\sum (\zeta_k(i, j)-\mu_{\zeta_k}(i,j))^2 \sum(\zeta^{ref}_k(i, j)-\mu_{\zeta_k^{ref}}(i,j))^2}} ,
        \label{eq:Elesim}
    \end{aligned}
\end{equation}
where $i,j$ denote the locations in height and width of the input features. $\mu_{\zeta_k(i,j)}$ is the mean of the feature on the second dimension. 
The element with a higher score means higher similar to the background. With a sigmoid function and a threshold value, $EleSim$ can be turned into the mask $M_k$ with the size of $(Bs, 1, H_k, W_k)$. Following the \cref{eq: disentanglement}, the disentangled representation is obtained.
We visualize some disentangled results in different layers of VGG-16 in \cref{fig: dis}. The disentangled results are growing better in deeper layers. It ensures that such disentangled representation can provide good patterns for following the processing step.

\subsection{Disentangled Representation}

The $EleSim$ operator can provide a good disentangled representation on a daytime domain $\mathcal{D}$, which benefits from the adequate training of VGG-16 and the distinguishing features of daytime images. However, these advantages are both absent in the nighttime. Few discriminative features in nighttime images lead to disastrous consequences in computing the similarity, especially in some scenes with extreme lightness conditions. 

For a generated daytime image $x_{\mathcal{N \rightarrow D}}$, the ultimate objective is becoming a realistic natural daytime image. In view of the good effectiveness on daytime images and the non-optimization nature of $EleSim$, the disentangling effectiveness of $EleSim$ is also a measurement for the generated quality.
We perform the $EleSim$ on the generated daytime to produce the mask $M_{\mathcal{N \rightarrow D}}$ and let it jointly optimize with the generated daytime image.
Reference to the $\zeta_k(x^{ref}_\mathcal{D})$, the optimization target is modeled as:


\begin{equation}
    \begin{aligned}
        \   \mathcal{L}_{back} &= \sum_k^{N} M_{\mathcal{N\rightarrow D}, k} \odot|| \ \zeta_k(x_\mathcal{N\rightarrow D}) - \zeta_k(x^{ref}_\mathcal{D})||_1,
	 \\
    \ & = \sum_k^{N}||\zeta_k^b(x_\mathcal{N\rightarrow D}) - \zeta_k^b(x^{ref}_\mathcal{D})||_1
    \label{eq: L_back_vgg}
    \end{aligned}
\end{equation}
where $k$ denotes the results in $k-th$ layer of VGG-16 $\zeta$. In \cref{eq: L_back_vgg}, the $M_{\mathcal{N\rightarrow D}, k}$ and $\zeta_k(x_\mathcal{N\rightarrow D})$ are updated together in each learning step. $M_\mathcal{N\rightarrow D}$ getting closer to the real daytime image, the mask $M_{\mathcal{N\rightarrow D}, k}$ will be convergent to the real mask of the original nighttime image $x_\mathcal{N}$.
Such strategy will force the generated image to be consistent with the domain-specific features of daytime, which will also push the mask of generated image $M_{\mathcal{N\rightarrow D}, k}$ produce a correct disentangled representation in the background regions.

Similar strategy is also applied in modeling the foreground. Chosen the generated daytime image to produce the mask, the second item loss function can be modeled as follow
\begin{equation}
    \begin{aligned}
        \ &  \zeta^f_k(x_\mathcal{N}) = (1 - M_{\mathcal{N\rightarrow D}, k}) \odot \zeta_k(x_\mathcal{N\rightarrow D}) \\
        \ & \zeta^f_k(x_{\mathcal{N \rightarrow D}}) = (1 - M_{\mathcal{N\rightarrow D}, k}) \odot \zeta_k(x_\mathcal{N \rightarrow D}), \\
        \ & \mathcal{L}_{fore} = d_f \, (\zeta^f(x_\mathcal{N}), \zeta^f(x_\mathcal{N\rightarrow D})).
    \end{aligned}	
    \label{eq: fore}
\end{equation}

The challenge is the modeling of distance measurement $d_i$.
Different from the \cref{eq: L_back_vgg}, the foreground features do not have the direct supervision from the daytime image $x_\mathcal{D} \in \mathcal{D}$. We have to employ the original nighttime image $x_\mathcal{N \rightarrow D}$ to provide the supervision. However, introducing the massive night domain information is adverse to our optimization objective. A distance measurement that model the abstract semantic rather the specific features, \emph{i.e.}: the texture and color, is urgent needed.


Thus, following the CUT \cite{park2020contrastive} and \cref{eq: optimization target}, we introduce our disentangled contrastive learning. 

The contrastive learning aims to provide a graph-like structure for the image by computing the similarity between any two elements. To construct a successful contrastive learning, two points are significant: the hard negative examples digging strategy and the sample quantity. Too few samples in computing the contrastive loss will have serious consequences in performance, which is common in the foreground modeling.
Based on our disentangled representation, we construct a contrastive learning strategy elaborated as follows. For each disentangled representation $M$, we will obtain a element-wise similarity scores before obtain the mask, which is denoted as $P$. $P$ presents the possibility of which each element is the foreground. By selecting the nearest neighbors around the 1 matrix, we can obtain the elements that is most similar to the foreground of generated image, which is also the hard negative examples for each other. Thus, we construct the hard negative examples and make sure the samples quantity will not be too small. Our contrastive loss is formulated as:
\begin{equation} \footnotesize
\begin{aligned}
    \ & \mathcal{L}_{fore} = \\
    &-\sum_k^{N} {log \frac{\exp(\zeta_k^f(x_\mathcal{N \rightarrow D})^T\zeta_k^f(x_\mathcal{N})/ \tau)}
    {\exp(\zeta_k^f(x_\mathcal{N \rightarrow D})^T\zeta_k^f(x_\mathcal{N})/ \tau)) + \exp(\zeta_k^b(x_\mathcal{N \rightarrow D})^T\zeta_k^b(x_\mathcal{N})/ \tau)}},
\end{aligned}	
\label{eq: our NCE}
\end{equation}
where $\tau$ is the hyper-parameter temperature. $k$ denotes the number of layer in VGG-16.
In practice, the mask of $\zeta_k^f$ is obtained from the generated image $x_\mathcal{N \rightarrow D}$. Such strategy makes sure that the disentangled representation can also focus on the graph structure. 
 

 Finally, we construct our merging operator with adversarial training so that we can model the merging operator into the parameters $w$ in $\mathcal{F}$.
 Following LSGAN\cite{Mao_2017_ICCV}, the adversarial loss is formulated as:
\begin{equation}
\begin{aligned}
    &\mathcal{L}_{adv}(\mathcal{F}) = ||D(x_\mathcal{N \rightarrow D})-1||_2^2,\\
    &\mathcal{L}_{adv}(D) = ||D(x_\mathcal{D})-1||_2^2 + ||D(x_\mathcal{N \rightarrow D})||_2^2,
\end{aligned}
\end{equation}
where $D$ denotes the discirminator network.

The final loss function is formatted as :

\begin{equation}
\begin{aligned}
    &\mathcal{L}_{total}(\mathcal{F}) = \mathcal{L}_{adv}(\mathcal{F}) + \mathcal{L}_{back} + \mathcal{L}_{fore},\\
    &\mathcal{L}_{total}(D) = \mathcal{L}_{adv}(D) .
\end{aligned}
\end{equation}

\begin{figure*}[htbp]
\centering
\includegraphics[width=1.0\linewidth]{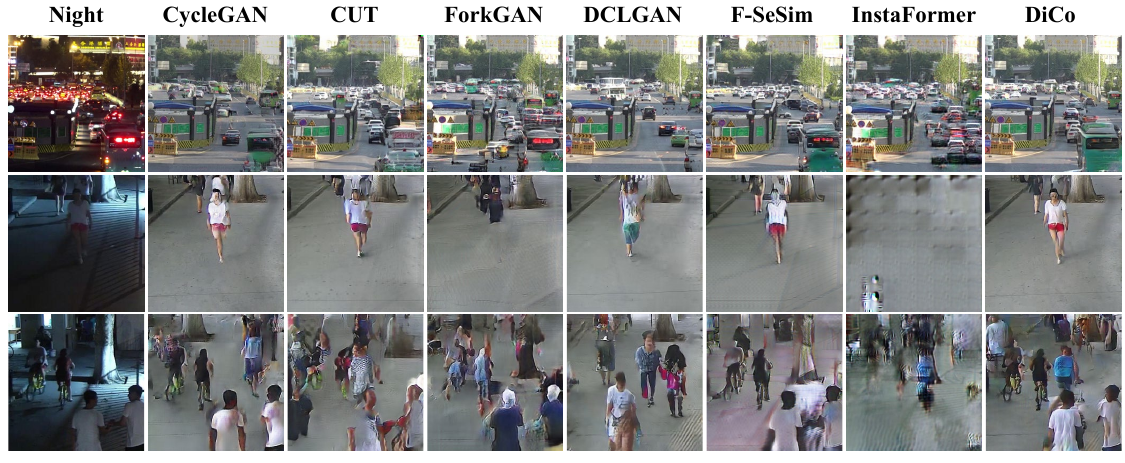} \
\vspace{-0.5cm}
\caption{The qualitative evaluation on \emph{NightSuR}. From top to down are: \textit{Traffic}, \textit{Pedestrian-g}, and \textit{Crowds}. }
\vspace{-0.2cm}
\label{fig:results2}
\end{figure*}

\begin{figure}[thbp]
\centering
\includegraphics[width=1.0\linewidth]{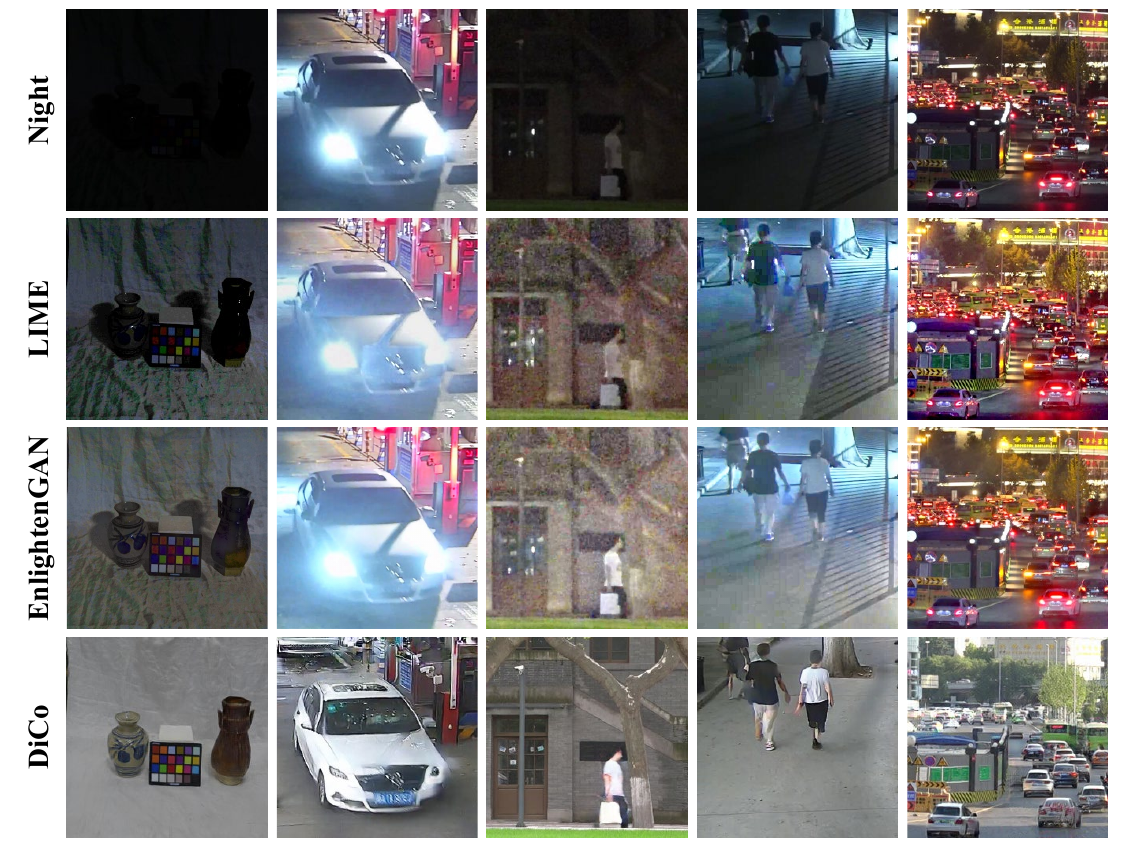} \
\vspace{-0.5cm}
\caption{The qualitative evaluation in comparison with low light enhancement methods }
\label{fig:results3}
\vspace{-0.2cm}
\end{figure}

\section{Experiments} \label{Sec: Experiments}

\subsection{Datasets}

To support our research on Night2Day for surveillance, a new dataset, NightSuR, is constructed in this paper. It contains 6 fixed scenes with 6574 images in total.

Firstly, we build an indoor dark scene (\textit{Darkness}) to simulate the night and day scenes in surveillance views. The light sources are controlled manually to simulate night and day. In Darkness, 167 pairs of images are collected in bright and dark environments. Secondly, two outdoor scenes (\textit{Traffic} and  \textit{Pedestrian-b}) are captured with fixed views of cameras.  \textit{Traffic} aims to simulate the data from the traffic monitor.  \textit{Pedestrian-b} presents the surveillance with tiny targets. Thirdly, three scenes from real surveillance cameras are selected (\textit{Pedestrian-g}, \textit{Crowds}, and \textit{Vehicle}). 

As discussed above, the background reference is the individual supervision for Night2Day in surveillance. Thus, an empty background image is captured from the time when the foreground is sparse. For the scenes that are impractical to capture a background image (\textit{Traffic}), we synthesize a background image by computing the average value across all the images in the scene.

Overall, our dataset covers widespread scenes and light conditions for surveillance, especially for some extreme environments (\emph{e.g.}, \textit{Vehicles} with \textit{flare}, and \textit{Darkness} with extremely low lightness). Additionally, it contains three scenes for pedestrians, which is common in surveillance and raises more challenges in detecting less orthogonal edges. The comparison of NightSuR and former datasets is demonstrated in \cref{tab:dataset} to conclude the necessity of the NightSuR in some significant characteristics. 

\setlength{\tabcolsep}{1pt}
\begin{table}[htbp]
\centering
\renewcommand\arraystretch{0.9}
\caption{The comparison of the proposed dataset and former datasets.}
\vspace{-0.2cm}

    \begin{tabularx}{0.5\textwidth}{X<{\centering} X<{\centering} X<{\centering} X<{\centering}X<{\centering}}
    \toprule
          &Color Night & Flare/Glow & Extreme Darkness & Daytime 	\\
   
    \hline

     TNO
      &few  & \XSolidBrush &\XSolidBrush &\Checkmark  \\
     INO	
     & few  &\XSolidBrush &\Checkmark &\Checkmark \\
     LLVIP 
     & \Checkmark  & few &\XSolidBrush & few \\
     NightSUR 
     &\Checkmark  &\Checkmark &\Checkmark &\Checkmark\\
    \bottomrule
    \end{tabularx}
\label{tab:dataset}
\vspace{-0.4cm}
\end{table}
\setlength{\tabcolsep}{1.4pt}

\subsection{Experimental Settings}
Experiments are conducted on the proposed dataset for Night2Day in surveillance scenes. Moreover, we also evaluate our method for driving scenes that are collected from the FLIR dataset to explore the extensibility of methods. 

\textbf{Evaluation Metric.} Following the common practice, we employ \emph{fr$\acute{e}$chet inception distance} (FID) scores \cite{heusel2017gans} to evaluate the quality of translated images. The FID scores results are shown in \cref{tab:1}.We compare DiCo with several effective image translation methods for qualitative comparison. Please refer to \cref{fig:results2} and \cref{fig:results3} for the reconstruction result.

\textbf{Implementation Details.} We implement our framework with a ResNet-based generator and PatchGAN-based discriminator. Our disentanglement module is based on the ImageNet-pretrained VGG-16, where the employed layers of VGG are $relu3-1$ and $relu4-1$. It should be noted that, in \textit{Darkness} dataset, the paired daytime data are not utilize to keep the unsupervised setting. Additionally, the results of compared methods are reproduced from their released source code.

\setlength{\tabcolsep}{1pt}
\begin{table*}[t]
\centering
\renewcommand\arraystretch{1.0}
\caption{The quantitative results. $\downarrow$ means the lower result is better.}
\vspace{-0.2cm}

    \begin{tabularx}{0.89\textwidth}{p{4cm}<{\centering} p{1.4cm}<{\centering} p{1.4cm}<{\centering} p{1.4cm}<{\centering} p{1.4cm}<{\centering}X<{\centering}X<{\centering} X<{\centering} X<{\centering} X<{\centering}}
    \toprule
     \multirow{2}{*}{Method}	&\multirow{2}{*}{\shortstack{ Cycle \\ Consistency}} &\multirow{2}{*}{sec/iter$\downarrow$}  &Traffic & Veh. & Crowd & Ped.-g & Ped.-b & Dark . & Ave.   	\\
     \cline{4-10}
     &&&\multicolumn{7}{c}{ FID$\downarrow$} \\
    \hline
    CycleGAN \cite{zhu2017unpaired}
    &\Checkmark & 0.135 & 62.8    & 89.6  &177.8  &97.1          & 51.6          & 219.2   & 116.4 \\

    CUT	\cite{park2020contrastive}	
    &\XSolidBrush & 0.086  & 81.0   & 72.3  &250.0  &123.6 & 64.0 & \underline{146.5} &122.9  \\
    NEGCUT \cite{wang2021instance}
    &\XSolidBrush & 0.111 &58.3&71.1&221.8&100.0&55.8&210.1&119.5 \\
    DCLGAN \cite{han2021dual}
    &\Checkmark & 0.228 &\bf{42.4}&\underline{70.1}&\underline{142.1}&105.1&59.5&154.8&\underline{95.7} \\
    
    F-LSeSim \cite{zheng2021spatially} 
    &\XSolidBrush & 0.047 &78.9 &259.7 &249 &201.5 &116.2 &278.3 &197.3 \\
    \hline
    ForkGAN \cite{zheng2020forkgan}
    &\Checkmark &  0.176   & 120.1&104.8 &236.1&188.0&143.8&334.1&187.8 \\
    InstaFormer \cite{kim2022instaformer}
    &\Checkmark & 0.503  &115.4&254.4&390.1&376.8&147.3&421.2&284.2 \\
    
    \hline
    LIME \cite{guo2016lime}	
    & - & - & 252.3  & 218.0 &269.5  &267.1         & 264.2         & 245.0   &252.7\\
    EnlightenGAN \cite{jiang2021enlightengan}
    &\XSolidBrush & - & 252.5 &211.9&251.5&222.5&255.9&222.1&236.1\\
    \hline
    \bf{DiCo}   	 
    &\XSolidBrush & 0.043  & \underline{44.2} & \bf{53.0}  & \bf{128.1} & \bf{75.8}   & \bf{32.3}   & \bf{82.8} & \bf{69.4}  \\
    DiCo w$\setminus$o LCI 
    & - & 0.029 &54.3&75.7&213.9&151.9&66.1&170.6&122.1  \\
    
    DiCo w$\setminus$o $L_{fore}$ 
    & - & - &60.2&76.9&222.4&108.1&\underline{42.7}&205.8&119.4  \\
    DiCo w$\setminus$o $L_{back}$ 
    & - & -    &53.8  &71.4 &172.1 &\underline{82.4} &139.7 &226.2 &122.1
    \\
    \bottomrule
    \end{tabularx}
\label{tab:1}

\end{table*}
\setlength{\tabcolsep}{1.4pt}

\begin{figure*}[htbp]
\centering
\includegraphics[width=1.0\linewidth]{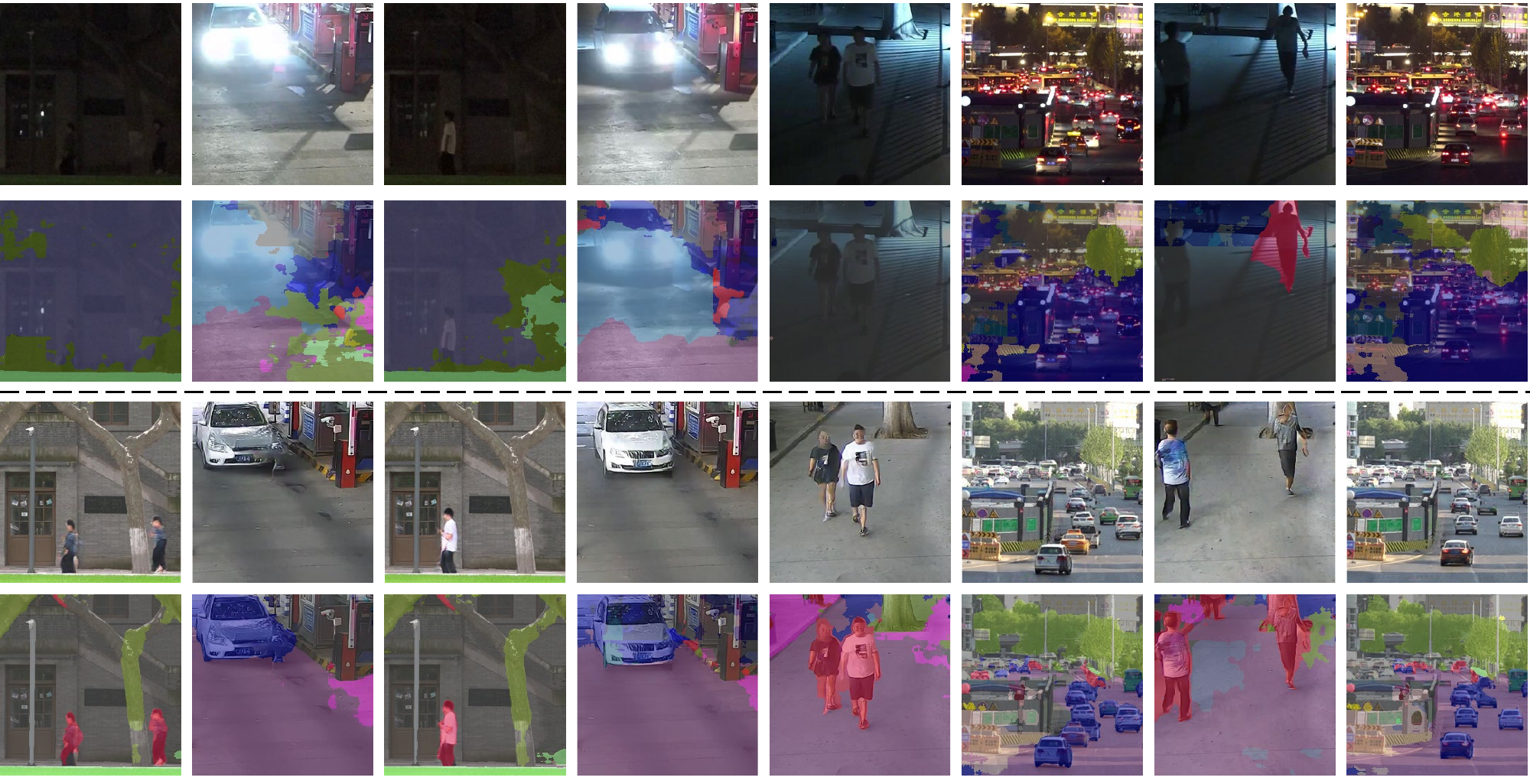} \
\vspace{-0.5cm}
\caption{The visualization of segmentation results. The first line and second line are original nighttime images and the corresponding semantic segmentation results. The third line and forth line are the synthetic daytime images by DiCo and the corresponding semantic segmentation results.}
\label{fig:seg_results}
\end{figure*}

\begin{figure*}[htbp]
\centering
\includegraphics[width=1.0\linewidth]{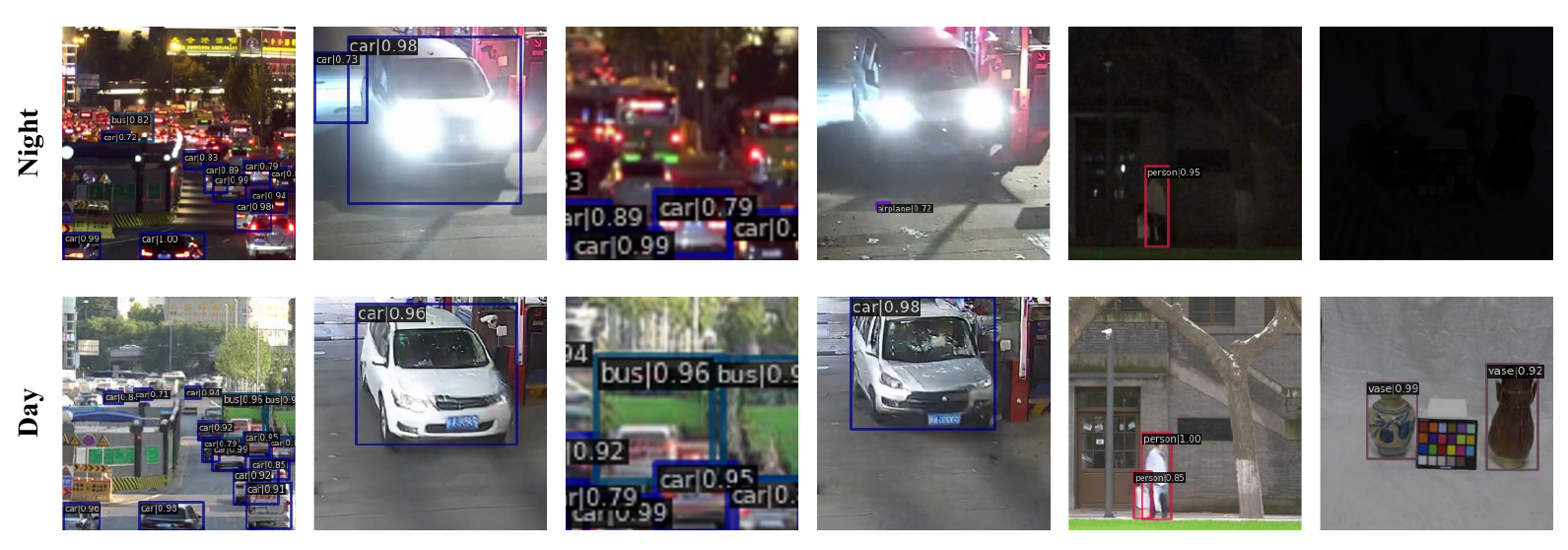} \
\vspace{-0.5cm}
\centering
\caption{The visualization of detection results in nighttime surveillance. The first line are detection results on original nighttime images and the second line are the results on synthetic daytime images by DiCo.}
\label{fig:det_results1}
\vspace{-0.2cm}
\end{figure*}

\subsection{Results on NightSuR}
In this paper, several unsupervised image-to-image translation methods are compared: CycleGAN \cite{zhu2017unpaired}, CUT \cite{park2020contrastive}, NEGCUT \cite{wang2021instance} and F-LSESim \cite{zheng2021spatially}. Some Night2Day methods for driving scenes, ForkGAN \cite{zheng2020forkgan} and InstaFormer \cite{kim2022instaformer}, are also in comparison. Note that InstaFormer \cite{kim2022instaformer} here is trained in unsupervised settings without a detection box as prior. We also compare two low-light enhancement methods to display the difference between Night2Day and low-light enhancement. 
In \cref{tab:1}, our method outperforms state-of-the-art among all the related works.
With the benefit of regression loss applied to the background, DiCo can easily outperform in the \textit{Pedestrian-b} in which the background is the major component. It could also explain the poor performance in Flir and suboptimal results in \textit{Traffic}.
\textit{Crowds} is crowded with little supervision of the background, which contributes to poor FID scores. Notably, despite the poor FID scores in \textit{Crowds} and \textit{Flir}, DiCo still achieves SOTA with the help of a learnable color invariant and disentangled contrastive learning strategy. This demonstrates the efficacy of our modules and reveals common issues of unsupervised image translation in complex scenes.
In addition, the DCLGAN \cite{han2021dual} is a strong baseline that reaches the suboptimal FID scores. It combines the strengths of cycle consistency training and contrastive learning, however, has its boundary under the design of traditional task formulation. 

The qualitative results are displayed in \cref{fig:results1} and \cref{fig:results2}. \cref{fig:results1} shows the results of some extremely dark and flaring scenes. The \cref{fig:results2} are some common application scenes.
Significantly, in the \textit{Vehicle}, DiCo can effectively manage flare using color invariant prior and disentangled contrastive learning, whereas others cannot. Moreover, due to the direct supervision from the background, DiCo can reconstruct the background with more details than others. 
Although F-SeSim \cite{zheng2021spatially} has a diminished capacity to reconstruct color information, it excels in semantic consistency protection for a variety of scenes. It motivates us to apply our method to additional scenes in future projects.
In supervised settings, InstaFormer \cite{kim2022instaformer} is a strong baseline because it is the only method with a vision transformer backbone. However, the vision transformer's potential remains untapped. Due to the data dependence of the vision transformer, more training data and steps may be helpful. This paper is, as far as we are aware, the first to successfully perform Night2Day on the human body, which lacks sufficient orthogonal edges like street scenes. Although the performance is not yet excellent, it demonstrates the possibility of achieving a higher standard of non-orthogonal Night2Day. 

\begin{table*}[t]
\centering
\renewcommand\arraystretch{1.0}
\caption{The FID results on BDD100k and FLIR1k.The lower result is better.}
\vspace{-0.2cm}

    \begin{tabularx}{0.89\textwidth}{X<{\centering} X<{\centering} X<{\centering} X<{\centering} X<{\centering}X<{\centering}X<{\centering} X<{\centering} X<{\centering} X<{\centering} X<{\centering}}
    \toprule
     & EnGAN & CycleGAN & DRIT &StarGAN &CUT & NEGCUT & F-LSeSim & ToDayGAN  & ForkGAN &DiCo \\	
    \hline
   BDD100k  &90.3  &51.7 &53.1 & 68.3  & -  & - & - & 43.8 & 37.6 & \bf{35.5}\\
  
    \hline
    FLIR1k &232.0 &178.2 & - & - & 214.3 & 306.5 &328.2 & - & 176.9 & \bf{153.2} \\

    \bottomrule
    
    \end{tabularx}
\label{tab:2}
\vspace{-0.4cm}
\end{table*}
\setlength{\tabcolsep}{1.4pt}

\begin{figure*}[htbp]
\centering
\includegraphics[width=1.0\linewidth]{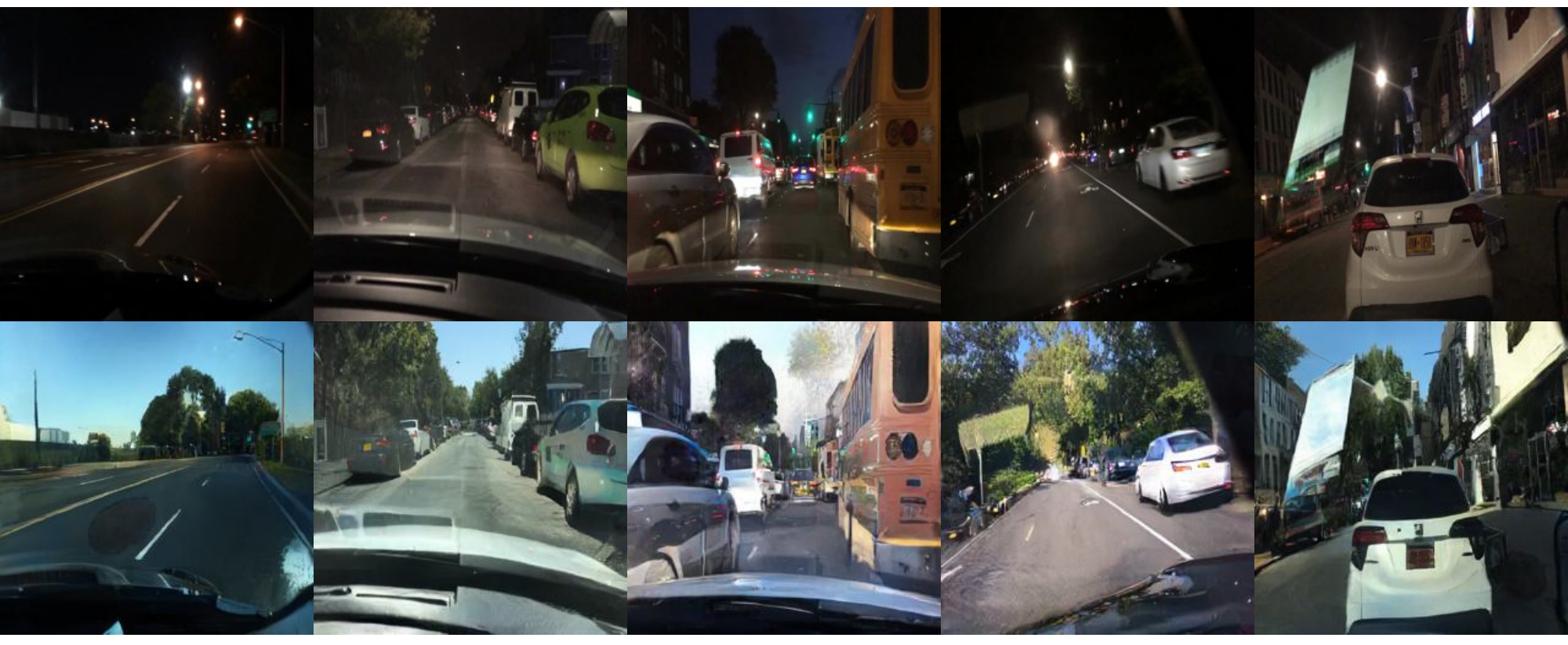} \
\caption{The qualitative evaluation on BDD100k dataset. The first line are the original nighttime images and the second line are the synthetic daytime images by DiCo.}
\label{fig:bdd100k}
\end{figure*}

\begin{figure}[htbp]
\centering
\includegraphics[width=1.0\linewidth]{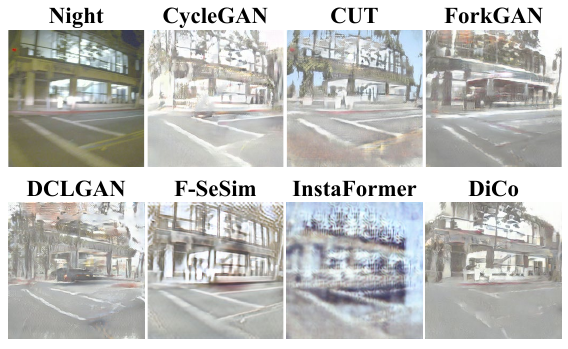} \
\caption{The qualitative evaluation on FLIR1k dataset}
\label{fig:flir}
\end{figure}
 
\subsection{Semantic Segmentation and Detection}

The semantic segmentation and detection experiments are also conducted to prove the ability in keeping the semantic consistency of DiCo. Specifically, we employ Faster-RCNN \cite{ren2015faster} as the detection model and Deeplabv3 \cite{chen2017rethinking} as the semantic segmentation model. The ResNet101 is employed as the backbone of Faster-RCNN and is trained on the COCO. The Deeplabv3 is trained on the Cityscapes dataset. We conduct inference with the two models on the generation results of DiCo and the nighttime images. The visualization of results is in \cref{fig:seg_results} and \cref{fig:det_results1}. 

DiCo mitigates the degradation of nighttime images in the following aspects. (1) It enhances some insignificant local patterns in nighttime images. For example, in the second line of \cref{fig:det_results1}, DiCo enhances the insignificant features of far-small objects in the nighttime (The buses and the children). (2) It corrects some confused local patterns at night. For example, in the second line of \cref{fig:det_results1}, DiCo mitigates the light on the ground that is similar to the headlight of vehicles, and finally corrects the detection results. (3) It adjusts the global features of nighttime images. For example, in \cref{fig:seg_results}, DiCo adjusts the global feature and corrects the pixel-wise classification, and achieves instance-aware results at night. 

\subsection{Expand to the Auto-Driving Scene}

Experiments in the driving scene datasets are also conducted to explore the expansion of DiCo. The FLIR and BDD100K are popular auto-driving datasets with sufficient daytime and nighttime images. 
For the FLIR dataset, we randomly chose 1000 pairs of nighttime and daytime images from the FLIR dataset to achieve a balanced distribution of nighttime and daytime images marked as FLIR1k. For the BDD100k, we choose clear daytime images and nighttime images in various kinds of weather to construct the Night2Day dataset.

We conduct DiCo in comparison with the former methods. The background in the driving scene is constructed with the average of a batch of daytime images. The quantitative results of FID scores are in the \cref{tab:2}. The qualitative results are in the \cref{fig:flir} and \cref{fig:bdd100k}. 
We can obtain two significant observations from the experiments. First, DiCo still shows expressive results in comparison with former work. It inspires us that disentangled representation is a general framework. It essentially provides the direct supervision for explicit invariance in target domain. In this work, we represent such explicit invariance in a simple way: the expectation of the target domain. It is within our expectation that improving such representation can provide even more surprising results.

 Second, it is obvious that the performance of DiCo is better on the BDD100k than the FLIR1k. It indicates that the perception of complex scenes still relies on the large-scale dataset. 
However, these experiments prove the expansion of DiCo and inspire us to develop disentangled representations with more generalized prior knowledge and solve problems in more general scenes.

\begin{figure*}[htbp]
\centering
\includegraphics[width=0.9\linewidth]{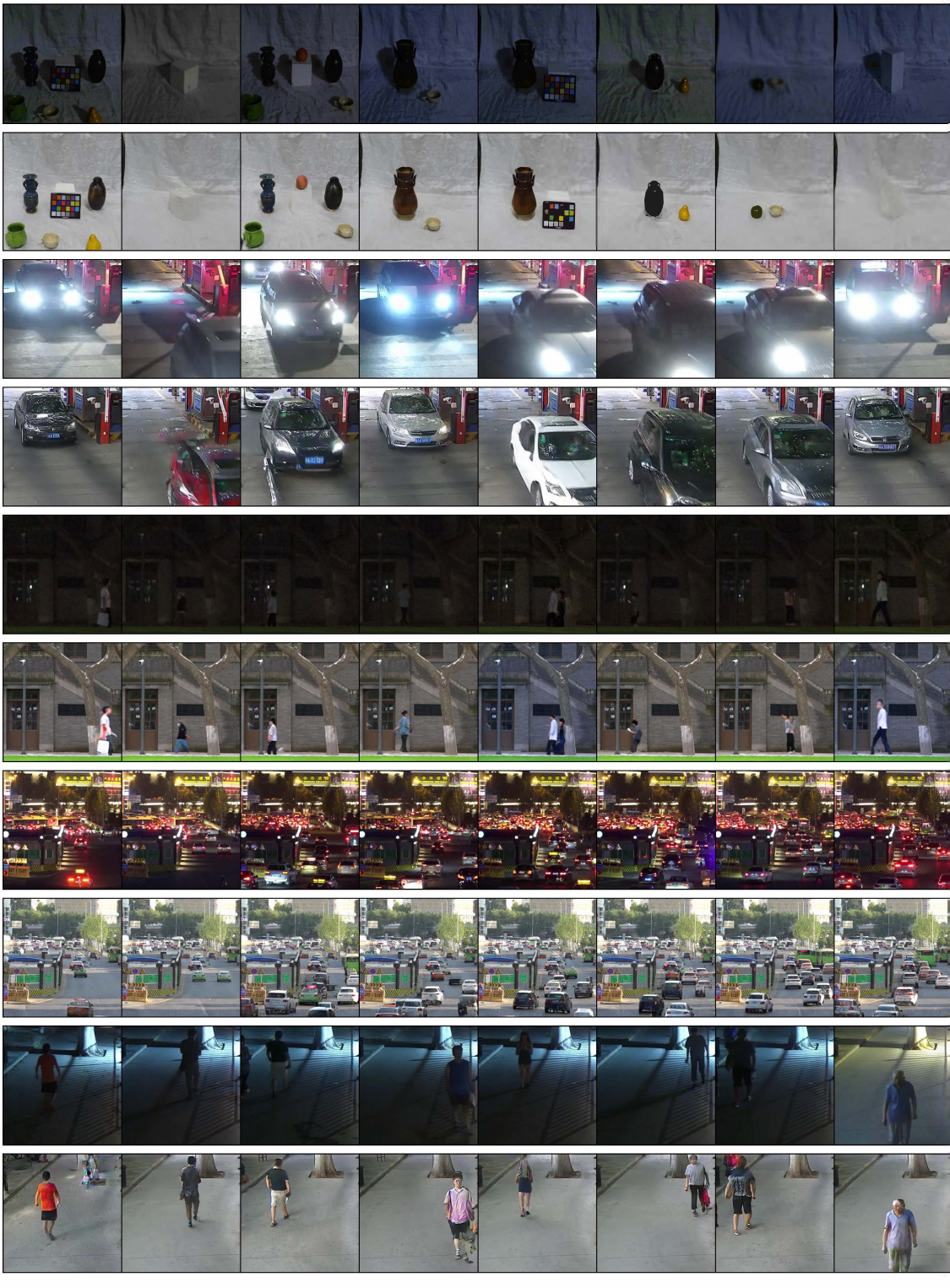}
\vspace{-0.4cm}
\caption{The auxiliary experimental results.}
\label{fig: more_results}
\end{figure*}

\subsection{Ablation Study}

\begin{figure}[htbp]
\centering
    \subfigure{
    \includegraphics[width=1.0\linewidth]{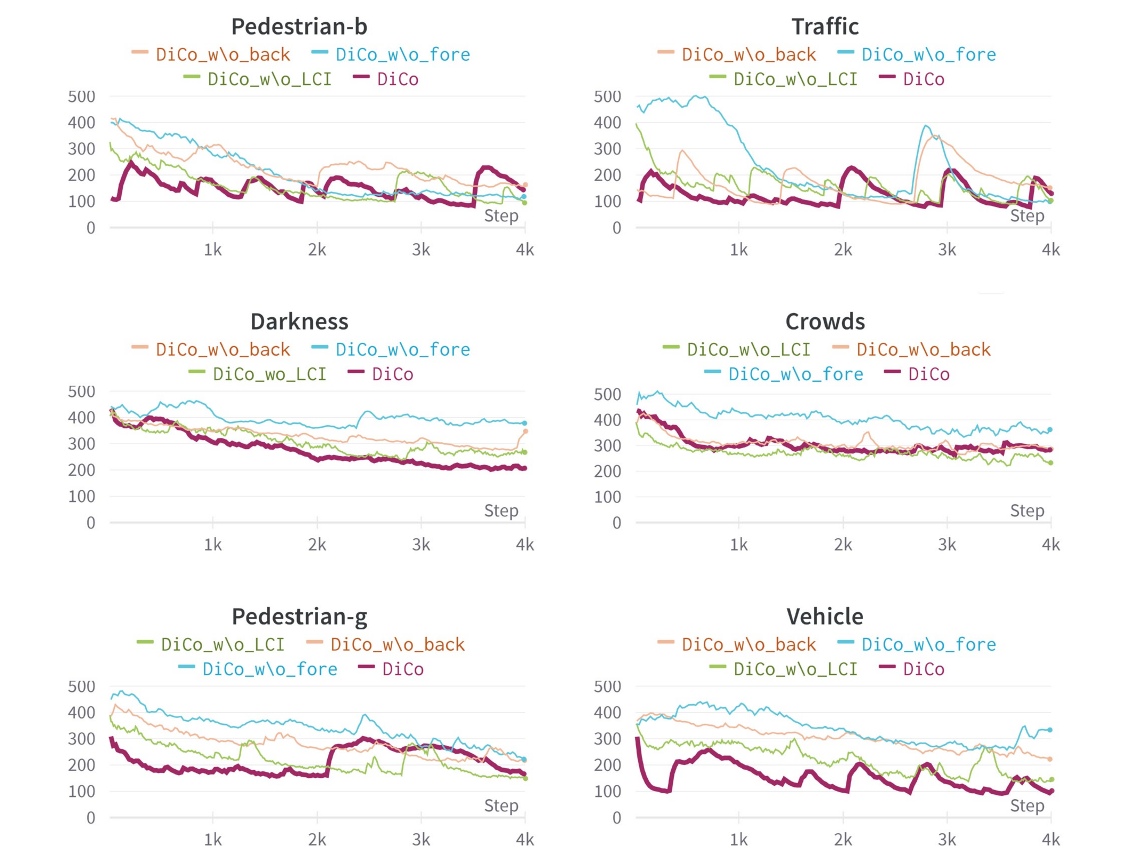} \
    }
\caption{The ablation FID scores during training process.  }

\label{fig:FID_curves}
\end{figure}
To figure out how these modules influence DiCo, our ablation study is conducted under these settings: DiCo w$\backslash$o LCI, DiCO w$\backslash$o $L_{fore}$, and DiCo w$\backslash$o $L_{back}$. Quantitative results are summarized in \cref{tab:1}.

The ablation of the three modules emphasizes different aspects of DiCo.
Under the DiCo w$\backslash$o LCI, scenes with gentle light conditions (\textit{Traffic} and \textit{Flir}) are less impacted while the dark scenes (\textit{Pedestrian-b}, \textit{Crowds}, and \textit{Darkness}) degrade the most. It confirms that the LCI module is important for accessing stable perception in various light conditions during the nighttime.
Surprisingly, the results in the \textit{Vehicle} with flare are less degraded. It shows that contrastive learning also contributes to the perception in flare scenes.

Under the DiCo w$\backslash$o $L_{fore}$ settings,  the model collapses in the driving scenes as shown in the column of \textit{Flir} in \cref{tab:1} and \cref{fig:flir}. In surveillance scenes, our model still performs well with the help of the disentangled representation. DiCo w$\backslash$o $L_{fore}$ performs suboptimal scores in \textit{Pedestrian-b}, which proves that our disentangled representation can control the semantic consistency effectively in surveillance, despite being limited by the reference background. The unexpected degradation in \textit{Darkness} indicates that the disentangled contrastive learning also contributes to the perception of the extreme environment. Such observation corresponds to the results of CUT \cite{park2020contrastive} and the ablation results of LCI in \textit{Vehicles}.

Under the DiCo w$\backslash$o $L_{back}$ setting, as shown in \cref{tab:1}, the performance of DiCo w$\backslash$o $L_{back}$ is closer to the CUT \cite{park2020contrastive}, but a bit better with our hard negative examples mining strategy and LCI module. As expected, DiCo degrades the most in the \textit{Pedestrian-b} and \textit{Darkness}, in which the background predominates the whole image.

Moreover, the FID scores during the training process are displayed in \cref{fig:FID_curves}. It shows the effectiveness of each module in DiCo, and presents that the whole DiCo model is more stable in training and faster in convergence, compared to other variants
.

\subsection{Limitations}
Although DiCo's performance is good in many surveillance scenes and has a good expansion in the driving scene. However, it still has some limitations. Firstly, it still results in some distortion in some small targets. Maybe high-resolution methods will help and they will be considered in our future work. Secondly, DiCo in the driving scene does not posse an apparent advantage over the former methods despite it being designed for surveillance scenes. The experiments on the driving scenes inspire us that disentangled representation is a general framework for describing the real world. Considering disentanglement in a more generalized view is our future work. Thirdly, Night2Day on complex scene still rely on the large-scale dataset. An apparent observation is that performance on BDD100k is distinctly better than the results on FLIR1k. We pursue to construct prior knowledge of the complex scenes and reduce dependencies on the large-scale dataset in our future work.


\section{Conclusion}
This paper presents a novel solution to nighttime surveillance. It aims to translate images from night to day while maintaining semantic consistency. To achieve this goal, we present the novel Disentangled Contrastive learning (DiCo) method, which primarily consists of a learnable color invariant, a disentangled representation, and a contrastive learning strategy. DiCo outperforms other strong baselines in traditional image translation, which shows the effectiveness of our method and reasonability of task formulation for nighttime surveillance. The excellent visual performance of DiCo also confirms that Night2Day is an effective solution to nighttime perception. In addition, a new Night2Day dataset for surveillance is created to support the research, which contains various surveillance scenes and light conditions. In the future, we will persistently explore perception in extreme conditions and expand the boundary of human cognition.

\ifCLASSOPTIONcaptionsoff
  \newpage
\fi





\bibliographystyle{IEEEtran}
\bibliography{IEEEabrv,Bibliography}

\vfill


\end{document}